\documentclass[11pt, a4paper]{article}

\usepackage[a4paper, top=2.5cm, bottom=2.5cm, left=2.5cm, right=2.5cm]{geometry}
\usepackage{amsmath, amssymb, amsfonts}
\usepackage{booktabs}
\usepackage{multirow}
\usepackage{authblk}
\usepackage{abstract}
\usepackage{hyperref}
\usepackage{cleveref}
\usepackage{enumitem}
\usepackage{graphicx}

\usepackage{mathptmx}
\usepackage{geometry}
\usepackage{authblk}
\usepackage{mathptmx}
\usepackage{microtype} 
\title{\textbf{Attention-Guided Flow-Matching \\ for Sparse 3D Geological Generation}}

\author[1,3$^{*}$]{Zhixiang Lu}
\author[2,3$^{*}$]{Mengqi Han}
\author[1]{Peixin Guo}
\author[1]{Tianming Bai}
\author[1]{Jionglong Su}
\author[3$^{\dag}$]{\\Fei Fang}
\author[1$^{\dag}$]{Sifan Song}
\affil[1]{Xi'an Jiaotong-Liverpool University, Jiangsu 215123, China}

\affil[2]{University of Chinese Academy of Sciences, Beijing 100049, China}

\affil[3]{Deep Optica, Shanghai 200030, China \authorcr 
\tt{fei.fang@deep-optica.com, Sifan.Song@xjtlu.edu.cn}}

\date{}

\begin{document}

\maketitle
\let\thefootnote\relax\footnotetext{$^{*}$ These authors contributed equally to this work.}
\let\thefootnote\relax\footnotetext{$^{\dag}$ Corresponding authors.}

\noindent\textbf{Abstract.} Constructing high-resolution 3D geological models from sparse 1D borehole and 2D surface data is a highly ill-posed inverse problem. Traditional heuristic and implicit modeling methods fundamentally fail to capture non-linear topological discontinuities under extreme sparsity, often yielding unrealistic artifacts. Furthermore, while deep generative architectures like Diffusion Models have revolutionized continuous domains, they suffer from severe representation collapse when conditioned on sparse categorical grids. To bridge this gap, we propose \textbf{3D-GeoFlow}, the first Attention-Guided Continuous Flow Matching framework tailored for sparse multimodal geological modeling. By reformulating discrete categorical generation as a simulation-free, continuous vector field regression optimized via Mean Squared Error, our model establishes stable, deterministic optimal transport paths. Crucially, we integrate 3D Attention Gates to dynamically propagate localized borehole features across the volumetric latent space, ensuring macroscopic structural coherence. To validate our framework, we curated a large-scale multimodal dataset comprising 2,200 procedurally generated 3D geological cases. Extensive out-of-distribution (OOD) evaluations demonstrate that 3D-GeoFlow achieves a paradigm shift, significantly outperforming heuristic interpolations and standard diffusion baselines.

\vspace{1.5em}


\section{Introduction}

Constructing high-resolution 3D geological block models from extremely sparse 1D borehole and 2D surface observation data is a highly ill-posed inverse problem, yet it remains fundamentally crucial for mineral exploration, geo-hazard assessment, and sustainable resource management \cite{wellmann2014geological, maccomack20193d}. The subsurface environment is inherently intricate, typically characterized by non-stationary spatial distributions, complex folded stratigraphy, multiphase faulting, and discontinuous intrusive bodies. 

Historically, this complex spatial interpolation problem has been addressed using simplistic deterministic and geostatistical methods. Heuristic techniques, such as 1D depth-wise interpolation, 2D polygonal methods and basic Kriging \cite{sinclair2002applied, oliver2015kriging}, have been widely utilized in the industry due to their computational tractability. Furthermore, implicit surface modeling methods have been developed to interpolate geological boundaries mathematically \cite{calcagno2008geological}. However, these conventional approximations fundamentally struggle to capture the non-linear geomorphological essence of complex strata. When confronted with extremely sparse conditioning data, they frequently produce unrealistic, staircase-like block models or excessively smooth boundaries that severely violate the fundamental principles of structural geology.

Recently, Deep Learning, particularly the emergence of advanced generative paradigms such as Generative Adversarial Networks (GANs) \cite{goodfellow2014generative}, Denoising Diffusion Probabilistic Models (DDPM) \cite{ho2020denoising}, and Continuous Flow-Matching \cite{lipman2022flow}, has shown tremendous promise in modeling complex, high-dimensional data distributions. In the computational geoscience domain, adapting these continuous-domain generative models to predict discrete, categorical geological grids presents profound challenges \cite{laloy2018training, zhang20233d}. Specifically, pure generative models suffer from severe performance degradation and representation collapse when conditioned on sparse exploration data (often representing less than 1\% of the target volume). Without explicit global spatial guidance, standard convolutional architectures fail to bridge the vast spatial void between distant borehole samples, leading to disjointed structural predictions.

To overcome this dilemma, we propose \textbf{3D-GeoFlow}, to the best of our knowledge, the first comprehensive framework to introduce Attention-Guided Continuous Flow-Matching for 3D multimodal geological modeling. To train and validate this framework robustly, we construct a novel, large-scale geological database comprising \textbf{2,200 high-fidelity 3D geological simulation cases}, encompassing multi-dimensional and multi-parameter variations (varying fault strikes, fold amplitudes, and intrusive dike dimensions). Our main contributions are summarized as follows:
\begin{itemize}[noitemsep, topsep=0pt]
    \item We pioneer the adaptation of Continuous Flow-Matching for categorical 3D geological volume generation under extreme sparsity. We mathematically justify and implement a Simulation-Free optimal transport formulation, replacing traditional cross-entropy with a highly stable Mean Squared Error (MSE) continuous vector field regression.
    \item We design an innovative \textit{Attention-Guided 3D U-Net} as the core neural ordinary differential equation (ODE) solver. By integrating custom 3D Attention Gates \cite{oktay2018attention}, the architecture effectively suppresses noisy background regions and forces the network to capture and propagate long-range spatial correlations from the sparse borehole and surface inputs across the entire 3D volume.
    \item We present a rigorous experimental evaluation using our newly curated dataset of 2,200 simulated cases. Extensive comparative analyses demonstrate that 3D-GeoFlow significantly outperforms traditional 1D/2D deterministic baselines and standard deep generative architectures in both voxel-wise precision and the structural realism of the generated geology.
\end{itemize}

\section{Related Work}
\label{sec:related}

The foundation of spatial interpolation and mineral inventory estimation relies heavily on deterministic and geostatistical frameworks. Early practices widely adopted 1D depth-wise interpolation and 2D polygonal methods due to their computational simplicity \cite{sinclair2002applied}. Subsequently, Kriging-based methods provided a mathematically rigorous foundation for spatial variance estimation \cite{isaaks1989applied, oliver2015kriging}. To model complex categorical patterns, Multiple-Point Geostatistics (MPS) was introduced to extract spatial statistics from training images \cite{strebelle2002conditional}. More recently, implicit modeling techniques utilizing potential-field interpolation and radial basis functions (RBF) have become the industry standard for constructing smooth geological interfaces \cite{calcagno2008geological}. However, these traditional methodologies exhibit severe limitations under extreme data sparsity. They heavily depend on stationary variogram assumptions or dense control points, invariably failing to automatically infer complex topological discontinuities and often yielding severe boundary artifacts or over-smoothed staircase block models. To alleviate the reliance on hand-crafted rules and stationary assumptions, Deep Learning, particularly 3D Convolutional Neural Networks (CNNs) and Generative Adversarial Networks (GANs) \cite{goodfellow2014generative}, has been introduced into computational geosciences. Laloy et al. \cite{laloy2018training} pioneered the use of spatial GANs for geostatistical inversion, while recent works have adapted conditional 3D-GANs to reconstruct geological models from drill-hole data \cite{zhang20233d, sui20213d}. Despite their ability to capture complex spatial priors, GAN-based approaches are notoriously difficult to train, frequently suffering from mode collapse. More importantly, standard 3D convolutions lack the global receptive field required to logically bridge the vast spatial voids between isolated borehole sequences, leading to disjointed strata representations when conditioned on extremely sparse inputs. To mitigate this, spatial attention mechanisms \cite{oktay2018attention, DeepGBTB2026, Lu2026attention} are critically needed to dynamically propagate localized structural features across the volumetric space.

\paragraph{Diffusion Models and Continuous Flow-Matching.} 
Denoising Diffusion Probabilistic Models (DDPM) \cite{ho2020denoising} and Score-based Stochastic Differential Equations (SDEs) \cite{song2020score, Lu2025prism} have profoundly revolutionized generative AI, surpassing GANs in distribution coverage and sample fidelity. Recently, Continuous Flow-Matching \cite{lipman2022flow, ghyselincks2026synthetic} emerged as a superior paradigm, bypassing the costly stochastic simulation of SDEs by explicitly learning an optimal transport vector field governed by Ordinary Differential Equations (ODEs). While these ODE/SDE frameworks are native to continuous domains, geological modeling requires predicting discrete categorical variables. Previous attempts to bridge this gap include discrete denoising diffusion (D3PM) \cite{austin2021structured}, which operates directly on categorical transition matrices. In this work, we take an orthogonal and more computationally efficient approach: we embed the discrete geological categories into a continuous metric space and formulate a Simulation-Free Flow-Matching objective. This allows us to leverage powerful continuous ODE solvers while utilizing hard physical masks to guarantee categorical exactness at sampled borehole locations.

\section{Methodology}

Due to the extreme sparsity of subsurface observations and the exorbitant cost of dense 3D geological drilling, training high-capacity deep generative models necessitates the use of high-fidelity synthetic datasets \cite{wellmann2014geological}. To minimize the domain gap between synthetic distributions and real-world geological environments, we developed a rigorous, multimodal data generation pipeline that couples kinematic geological modeling with computational geophysics. 

\subsubsection{Procedural 3D Geological Modeling}
We utilize a procedural geological engine inspired by kinematic and implicit modeling paradigms \cite{delavarga2019gempy} to generate diverse, geologically plausible 3D volumes. The target domain is discretized into a $64 \times 64 \times 64$ voxel grid. To simulate complex stratigraphy and structural evolution, we model a continuous sequence of geological events modeled as a Markov geostory. 

Specifically, the baseline geological history incorporates sequential deposition of basement and host sedimentary sequences, regional tilting, and multi-phase tectonic deformations including plunging folds and normal faulting. Furthermore, to mimic economically significant hydrothermal alteration systems, we explicitly model intrusive dike planes and progressive alteration halos (e.g., propylitic, argillic, and phyllic-silicic zones). Through combinatorial permutations of structural parameters and random noise perturbations, we systematically generate a highly diverse dataset comprising 9 distinct lithological and alteration facies.
\begin{figure}[t]
  \centering
 \includegraphics[width=.95\textwidth]{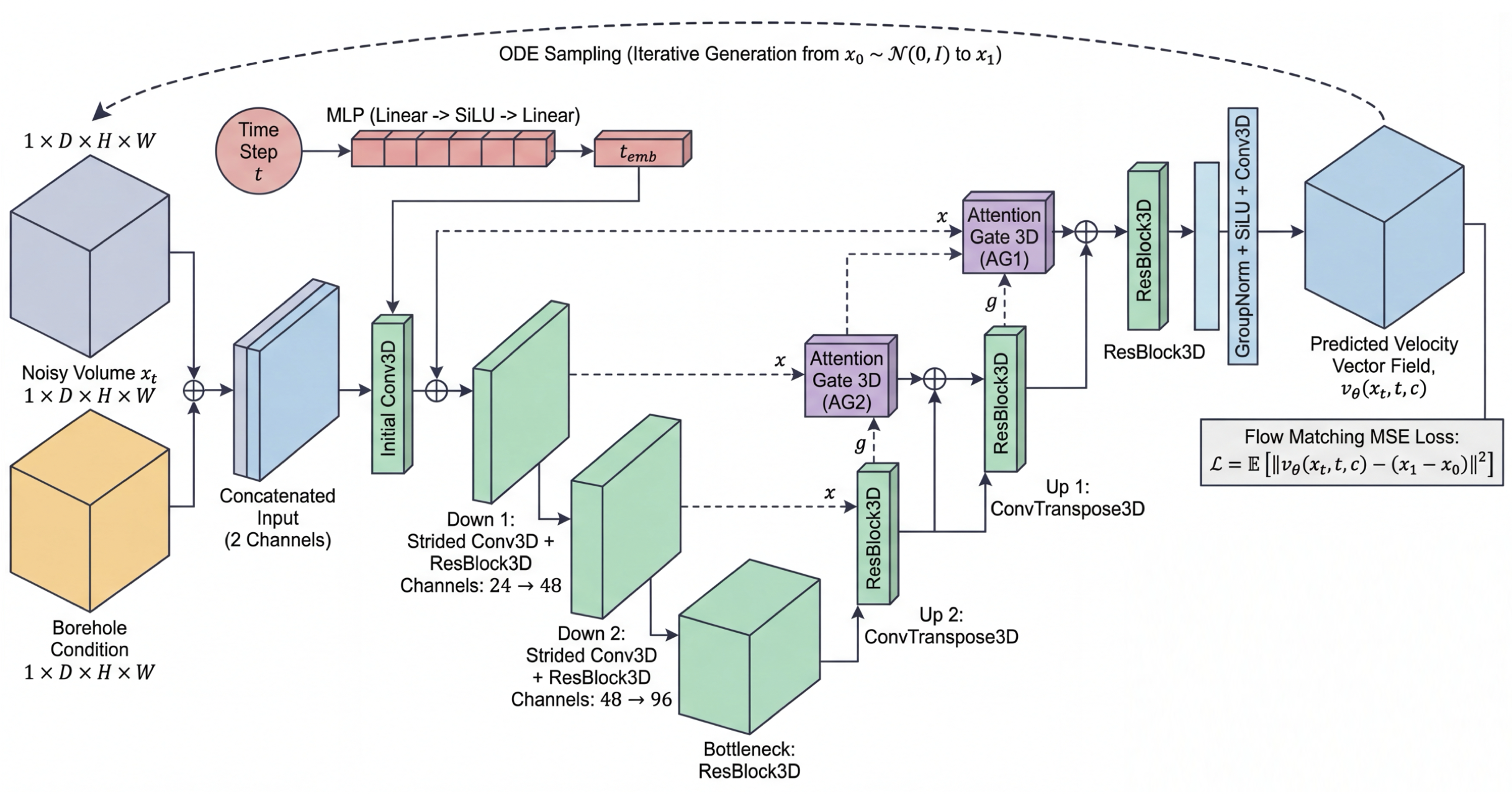}
  \caption{The architecture of Attention-Guided Continuous Flow-Matching (3D-GeoFlow) model.}
  \label{fig:framework}
\end{figure}
\label{sec:method}
\subsection{Dataset Construction}
\label{sec:dataset}
\subsubsection{Petrophysical Mapping and Forward Simulation}
To provide multimodal conditioning data, we synthesize the corresponding geophysical potential fields (gravity and magnetics). First, the discrete 3D lithological models are mapped to continuous petrophysical property grids. Density contrasts (relative to a $2050 \text{ kg/m}^3$ background) and magnetic susceptibility contrasts (relative to a $5.0 \times 10^{-4}$ SI background) are assigned to each of the 9 categories based on standard empirical rock physics bounds.

The 3D forward simulations are executed using the \textit{SimPEG} (Simulation and Parameter Estimation in Geophysics) framework \cite{cockett2015simpeg}. We formulate the gravity and magnetic responses as a 3D integral equation over the active computational mesh. For the magnetic simulation, a uniform inducing background field is applied with a designated amplitude ($50,000$ nT), inclination ($-50^\circ$), and declination ($10^\circ$). The synthetic responses are recorded at a $30 \times 30$ receiver grid draped over the target area. To simulate realistic survey conditions and sensor imperfections, zero-mean Gaussian noise with a standard deviation $\sigma \sim \mathcal{U}(0.005, 0.01)$ is superimposed onto the theoretical forward responses.

\subsubsection{Sparse Sampling and Masking}
To replicate real-world exploration constraints, we extract sparse representations from the dense 3D volumes. A surface-and-borehole mask is generated by randomly sampling locations along the $X-Y$ plane and extracting vertical 1D traces (simulating drill holes) down the $Z$-axis. Unsampled regions are masked with a distinct token ($-1$), while the air/terrain boundary is assigned a dedicated label. This yields the highly sparse condition tensor $c$, paired with the dense ground-truth volume $x$ and the 2D geophysical arrays, forming the complete multimodal tuple for network training.

\subsection{Denoising Diffusion Probabilistic Models}
In contrast to the explicit vector field regression of Flow-Matching, Denoising Diffusion Probabilistic Models (DDPMs) \cite{ho2020denoising} conceptualize generation as a stochastic Markovian denoising process. Within this framework, the objective diverges from discrete classification; instead, the neural network functions as a \textit{Continuous Noise Estimator} or, equivalently, a score function approximator \cite{song2020score}.

Let $x_0$ denote the ground truth 3D geological volume, mapped into a continuous metric space, and $c$ represent the sparse surface and borehole condition. The forward diffusion process gradually corrupts $x_0$ into pure Gaussian noise $x_T \sim \mathcal{N}(0, I)$ over time $t \in [0, T]$. Thanks to the reparameterization trick, the perturbed state $x_t$ at any arbitrary timestep $t$ can be sampled in closed form:
\begin{equation}
    x_t = \sqrt{\bar{\alpha}_t} x_0 + \sqrt{1 - \bar{\alpha}_t} \epsilon
\end{equation}
where $\epsilon \sim \mathcal{N}(0, I)$ is the randomly sampled standard Gaussian noise, and $\bar{\alpha}_t$ represents the cumulative variance schedule.

The generative (reverse) process aims to iteratively denoise $x_T$ back to the empirical data distribution $x_0$. To achieve this, the neural network $\epsilon_\theta(x_t, t, c)$ is trained to predict the exact noise vector $\epsilon$ that was injected at timestep $t$. Because this involves regressing a continuous perturbation trajectory, we strictly optimize the network using the Mean Squared Error (MSE) Loss:
\begin{equation}
    \mathcal{L}(\theta) = \mathbb{E}_{t, x_0, \epsilon} \left[ \left\| \epsilon_\theta(x_t, t, c) - \epsilon \right\|^2 \right]
\end{equation}
Minimizing this L2 norm is mathematically profound: it is equivalent to optimizing a reweighted variational lower bound (ELBO) on the data likelihood, and intrinsically matches the Stein score (the gradient of the log-density) of the perturbed data distribution \cite{ghyselincks2026synthetic}. Upon successful MSE optimization, we can employ an ancestral sampler or a reverse Stochastic Differential Equation (SDE) solver during inference, stochastically stepping backward from pure noise to stably generate high-fidelity 3D geological structures.

\subsection{Continuous Flow-Matching}
In standard semantic segmentation tasks, models predict discrete categories utilizing Cross-Entropy (CE) loss \cite{Causalsamllm, Lu2026skinclipvl}. However, in Continuous Flow-Matching and Diffusion models, the objective diverges fundamentally: the model is not acting as a direct classifier, but rather as a \textit{Continuous Vector Field Regressor}.

Let $x_1$ denote the ground truth 3D geological volume, mapped into a continuous metric space, and $x_0 \sim \mathcal{N}(0, I)$ denote pure Gaussian noise. The generative process is governed by an Ordinary Differential Equation (ODE):
\begin{equation}
    d x_t = v_\theta(x_t, t, c) dt
\end{equation}
where $v_\theta$ is our neural network (the velocity field), $t \in [0, 1]$ is the time step, and $c$ represents the sparse surface and borehole condition.

Our goal is to force the neural network $v_\theta(x_t)$ to approximate the target velocity field $u_t = x_1 - x_0$, which geometrically points from the pure noise distribution to the true data distribution. Because this is a continuous regression problem on a probability path, we strictly employ the Mean Squared Error (MSE) Loss:
\begin{equation}
    \mathcal{L}(\theta) = \mathbb{E}_{t, x_0, x_1} \left[ \left\| v_\theta(x_t, t, c) - (x_1 - x_0) \right\|^2 \right]
\end{equation}
Using the L2 norm (MSE) to minimize the discrepancy between the predicted and actual velocity vectors provides mathematical guarantees that the model learns the most efficient Probability Density Path (Optimal Transport). Upon successful MSE optimization, we can employ an ODE solver during inference to integrate along this vector field, stably generating the target 3D geological structures.

In conventional 3D U-Net architectures, skip connections uniformly concatenate low-level encoder features with high-level decoder representations. While effective for dense volumetric data, this paradigm catastrophically fails under extreme geological sparsity. The encoder extracts local features predominantly consisting of unconditioned noise (representing empty, unexplored subsurface regions). Directly concatenating this unfiltered noise corrupts the generative vector field $v_\theta$, leading to severe topological artifacts.
\begin{figure}[t]
  \centering
 \includegraphics[width=\textwidth]{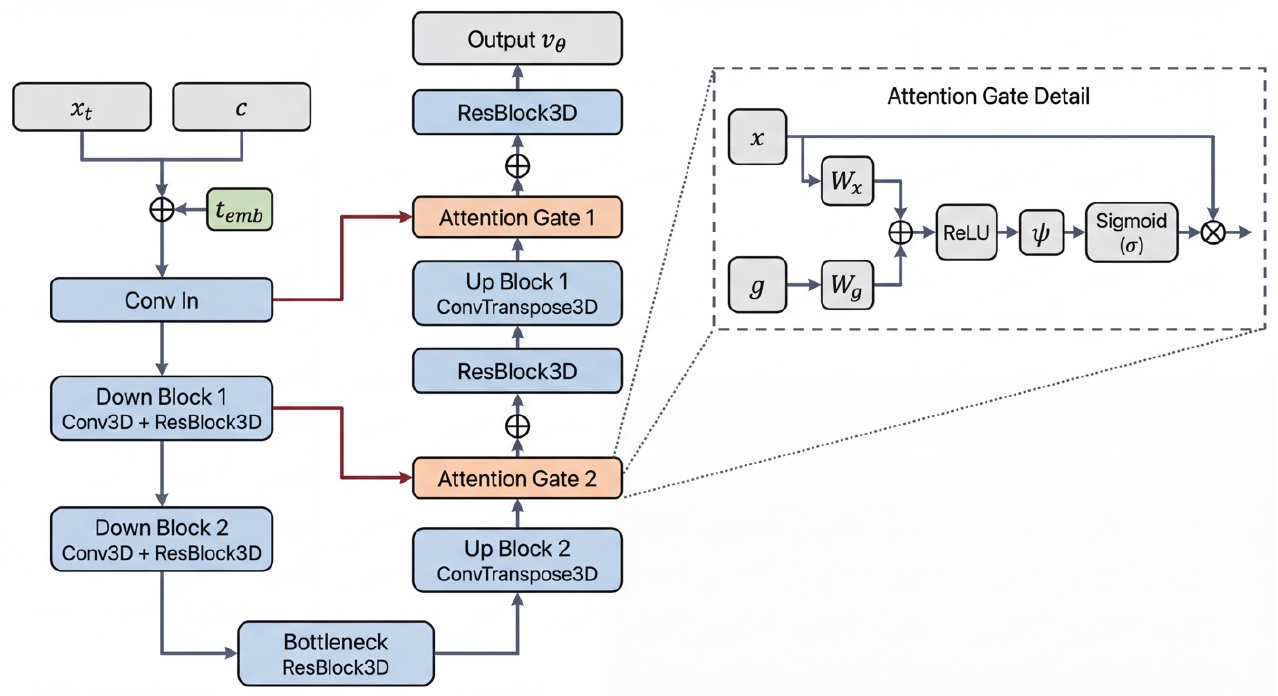}
  \caption{Detailed architecture of the 3D Spatial Attention Gates (3D-SAG). This mechanism spatially filters skip-connection features, compelling the model to focus on the long-range structural continuity implied by the sparse borehole conditions while suppressing unexplored noise regions.}
  \label{fig:attention}
\end{figure}
\subsection{3D Spatial Attention for Sparse Conditioning}
\label{sec:attention}
To address this ``noise leakage'' problem, we introduce a \textbf{3D Spatial Attention Gate (3D-SAG)} explicitly designed to dynamically filter the skip-connection pathways. The core intuition is to utilize the deep, semantically rich gating signal from the decoder to spatially evaluate and activate only the structurally relevant regions of the shallow encoder features.

Let $x_l \in \mathbb{R}^{F_l \times D \times H \times W}$ denote the local feature map extracted from the encoder at layer $l$, and $g \in \mathbb{R}^{F_g \times D \times H \times W}$ denote the gating signal from the corresponding coarse level of the decoder. We project both tensors into a shared intermediate representation space $F_{int}$ via linear transformations $\mathbf{W}_x$ and $\mathbf{W}_g$. The voxel-wise attention coefficients $\alpha \in [0, 1]^{1 \times D \times H \times W}$ are formulated as:

\begin{equation}
    q_{att} = \sigma_1 \left( \mathcal{GN}\left( \mathbf{W}_x \ast x_l \right) \oplus \mathcal{GN}\left( \mathbf{W}_g \ast g \right) \right)
\end{equation}
\begin{equation}
    \alpha = \sigma_2 \left( \mathcal{GN}\left( \mathbf{W}_\psi \ast q_{att} \right) \right)
\end{equation}
\begin{equation}
    \hat{x}_l = x_l \odot \alpha
\end{equation}

where $\ast$ denotes a $1 \times 1 \times 1$ 3D convolution, $\oplus$ denotes element-wise addition, and $\odot$ represents the Hadamard (element-wise) product. 

To ensure stable gradients across deep 3D networks, particularly when constrained by small batch sizes due to GPU memory limitations in volumetric generation, we rigorously employ Group Normalization ($\mathcal{GN}$) rather than standard Batch Normalization. The non-linear activation functions $\sigma_1(x) = \max(0, x)$ (ReLU) and $\sigma_2(x) = \frac{1}{1 + e^{-x}}$ (Sigmoid) constraint the attention maps into a strict probability range $[0, 1]$. By deploying $\hat{x}_l$ instead of the raw $x_l$ into the subsequent decoder layers, the model actively suppresses the unconditioned background void, forcing the generative vector field to prioritize the long-range geological continuity implied by the sparse borehole anchors.

\begin{table}[t!]
    \centering
    \caption{Out-Of-Distribution Performance Comparison of 3D Geological Modeling Methods.}
    \label{tab:results}
    \begin{tabular}{l c c c}
        \toprule
        \textbf{Method} & \textbf{Acc (Incl. Air)} & \textbf{Acc (Excl. Air)} & \textbf{mIoU (Excl. Air)} \\
        \midrule
        \multicolumn{4}{l}{\textit{Rule-based Deterministic Baselines}} \\
        Depth-wise Majority \cite{sinclair2002applied} & 52.81\% & 17.55\% & 5.36\% \\
        Polygonal Method \cite{wellmann2014geological} & 53.07\% & 18.00\% & 5.73\% \\
        \midrule
        \multicolumn{4}{l}{\textit{Standard Deep Generative Baselines}} \\
        Diffusion (DDPM) \cite{ho2020denoising} & 48.82\% & 10.57\% & 4.65\% \\
        Flow-Matching \cite{lipman2022flow}& 74.17\% & 54.87\% & 27.51\% \\
        \midrule
        \textbf{3D-GeoFlow (Proposed)} & \textbf{83.50\%} & \textbf{71.17\%} & \textbf{34.46\%} \\
        \bottomrule
    \end{tabular}
\end{table}
\section{Experiments and Results}
\label{sec:experiments}

\subsection{Experimental Setup}
To rigorously evaluate the generalization capabilities of the generative models, we established a stringent Out-Of-Distribution (OOD) testing protocol. The OOD dataset comprises an entirely unseen geological region containing 52,317,936 voxels distributed across 9 geological and alteration classes. To simulate real-world exploration bottlenecks, the sole input conditions provided to all models during inference are strictly limited to the 2D surface topography and highly sparse 1D borehole traces.
We benchmark our proposed \textbf{3D-GeoFlow} against two categories of baselines: 
1) \textit{Traditional Rule-based Methods}: 1D Depth-wise Majority and 2D Polygonal (Nearest-Neighbor) interpolation \cite{sinclair2002applied, wellmann2014geological}.
2) \textit{Deep Generative Models}: Pure DDPM (Denoising Diffusion Probabilistic Models) \cite{ho2020denoising} and the Standard Continuous Flow Matching \cite{lipman2022flow}.

\subsection{Quantitative Results and Analysis}

\begin{table}[b]
    \centering
    \caption{\textbf{Quantitative Evaluation of Per-Class Performance.} We report the Accuracy (Acc) and Intersection over Union (IoU) for the geological classes. Best results are highlighted in \textbf{bold}.}
    \label{tab:per_class_metrics_with_dist}
    \resizebox{\linewidth}{!}{
    \begin{tabular}{l c cc cc}
        \toprule
        \multirow{2}{*}{\textbf{Geological Class}} & 
        \multirow{2}{*}{\textbf{Proportion (\%)}} & 
        \multicolumn{2}{c}{\textbf{Baseline (Standard FM)}} & 
        \multicolumn{2}{c}{\textbf{3D-GeoFlow (Attention-Guided)}} \\
        \cmidrule(lr){3-4} \cmidrule(lr){5-6}
        & & \textbf{Acc (\%)} & \textbf{IoU (\%)} & \textbf{Acc (\%)} & \textbf{IoU (\%)} \\
        \midrule
        01 Air                       & 42.77 & 100.00 & 99.99 & \textbf{100.00} & \textbf{100.00} \\
        04 Mt Janet Andesite         & 19.76 & 53.66 & 45.01 & \textbf{84.67} & \textbf{70.08} \\
        02 Molly Darling Sandstone   & 17.33 & 64.37 & 60.35 & \textbf{81.93} & \textbf{75.81} \\
        07 Surface Sand / Soil       & 7.66  & \textbf{61.66} & 54.85 & 59.05 & \textbf{56.50} \\
        03 Ignimbrite                & 5.73  & 45.97 & 20.85 & \textbf{49.58} & \textbf{29.80} \\
        05 Conglomerate              & 4.19  & 48.13 & 18.12 & \textbf{54.88} & \textbf{32.92} \\
        06 Siltstone / Mudstone      & 1.77  & \textbf{10.59} & \textbf{3.13}  & 6.24  & 1.98  \\
        08 Outer Argillic Alteration & 0.46  & \textbf{7.44}  & \textbf{6.36}  & 4.09  & 3.92  \\
        09 Phyllic + Silicification  & 0.34  & \textbf{16.54} & \textbf{11.38} & 4.80  & 4.64  \\
        \midrule
        \textbf{Overall (Excl. Air)} & 100     & 54.87 & 27.51 & \textbf{71.17} & \textbf{34.46} \\
        \bottomrule
    \end{tabular}
    }
\end{table}
\paragraph{Overall OOD Performance.}
The comprehensive quantitative metrics on the OOD test set are summarized in \Cref{tab:results}. Traditional rule-based baselines perform poorly (achieving roughly 18\% subsurface accuracy), as they are mathematically incapable of inferring non-linear geological topologies from sparse discrete points. Surprisingly, the standard DDPM architecture fails catastrophically in this sparse 3D categorical setting (10.57\% subsurface accuracy). This degradation occurs because the stochastic Markovian noise injection inherently destroys structural coherence when the spatial conditioning is extremely weak. In contrast, the Standard Flow Matching baseline establishes a solid foundation (54.87\%) due to its deterministic, straight-path ODE sampling. However, our proposed \textbf{3D-GeoFlow (Attention-Guided FM)} achieves a massive leap, pushing the subsurface accuracy to \textbf{71.17\%} and the mIoU to \textbf{34.46\%}. This demonstrates that the integrated 3D Attention Gates successfully extract and propagate global spatial correlations from the sparse borehole hints across the entire volumetric space.

\begin{figure}[t!]
  \centering
  \includegraphics[width=\textwidth]{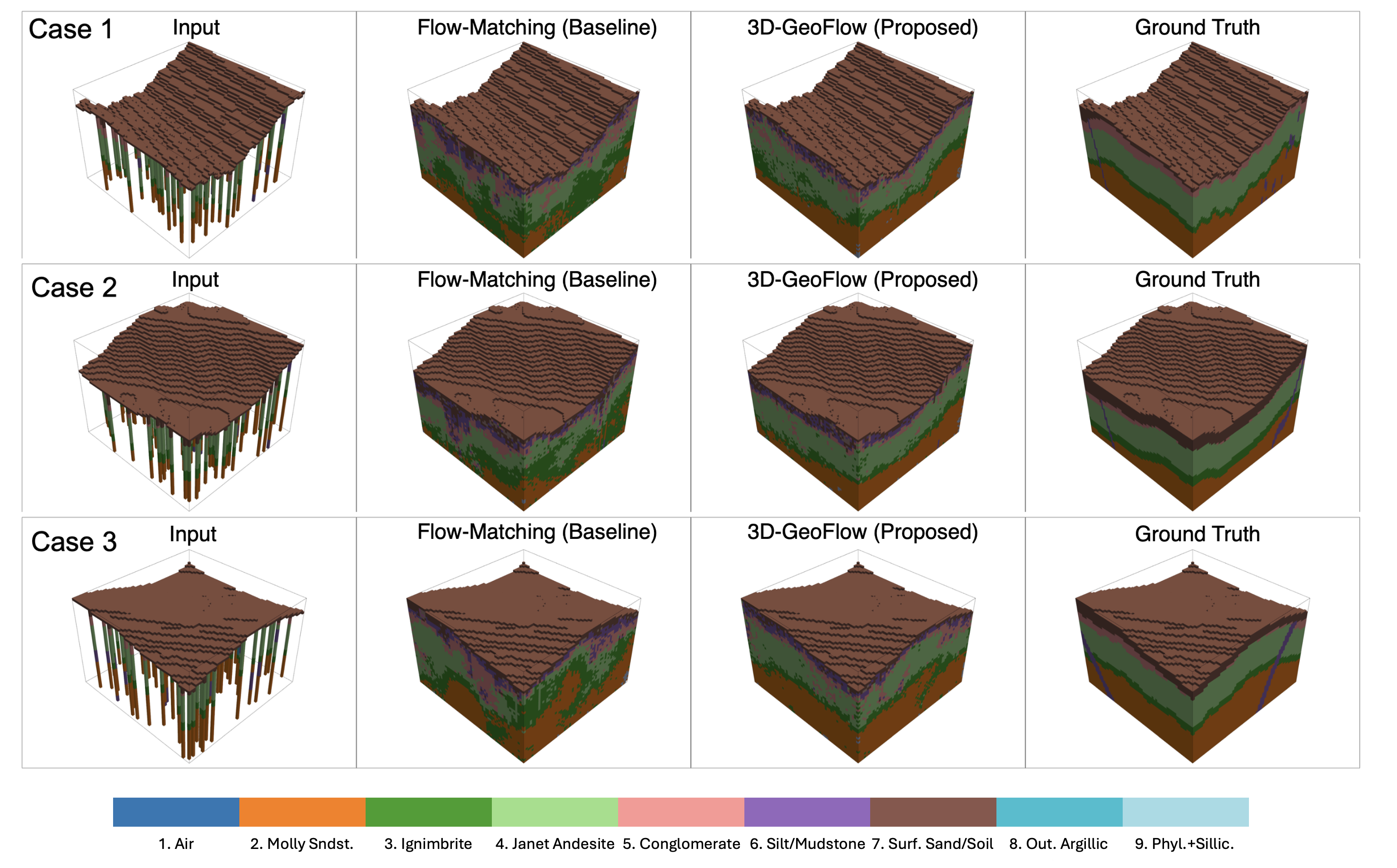}
  \caption{Qualitative 3D rendering results on the Out-Of-Distribution (OOD) test set.}
  \label{fig:rendering}
\end{figure}

\begin{figure}[htbp]
  \centering
  \includegraphics[width=.9\textwidth]{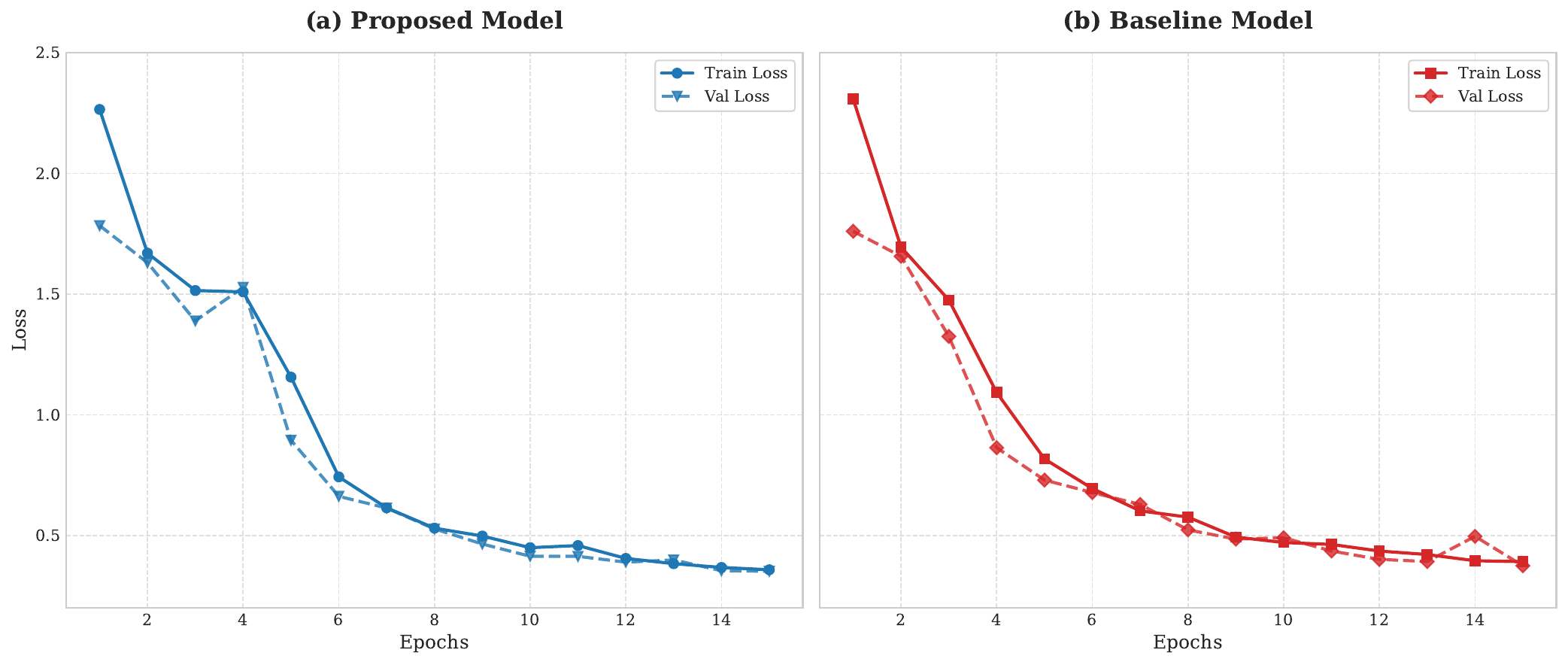}
  \caption{Training and validation loss curves over epochs between 3D-GeoFlow (Proposed) and flow-matching (Baseline) model.}
  \label{fig:loss}
\end{figure}

\paragraph{Per-Class Analysis and The Long-Tail Challenge.}
To further dissect the model's performance, \Cref{tab:per_class_metrics_with_dist} details the voxel-wise Accuracy and IoU for the 9 classes, sorted by their volumetric proportions. The dataset exhibits a severe long-tail distribution, common in natural geosciences, where dominant host rocks (e.g., Mt Janet Andesite, Molly Darling Sandstone) constitute the vast majority of the subsurface volume, while hydrothermal alteration zones occupy less than 1\%.

Our 3D-GeoFlow architecture shows overwhelming dominance in reconstructing the primary stratigraphic framework. For instance, the accuracy for \textit{Mt Janet Andesite} surges from 53.66\% to \textbf{84.67\%}, and \textit{Molly Darling Sandstone} improves from 64.37\% to \textbf{81.93\%}. This confirms that the attention mechanism perfectly captures the macroscopic structural geology. 

We do observe a slight performance trade-off in the extreme long-tail classes. This is a well-documented phenomenon in volumetric deep learning under extreme sparsity: to minimize the global energy landscape during the vector field regression, the attention mechanism inherently prioritizes the macroscopic continuity of dominant bounding strata. Consequently, highly localized, micro-scale alteration halos may be overly smoothed unless directly intersected by a conditioning borehole. Despite this, the substantial gains in the macroscopic structural topology render 3D-GeoFlow vastly superior for regional 3D geological modeling.

\section{Conclusion}
\label{sec:conclusion}
In this paper, we introduced 3D-GeoFlow, a novel Attention-Guided Continuous Flow Matching framework engineered to solve the highly ill-posed inverse problem of 3D geological modeling under extreme data sparsity. By elegantly formulating categorical voxel generation as a continuous vector field regression optimized via Mean Squared Error, we bypassed the computational and stochastic bottlenecks of traditional diffusion models. Furthermore, by integrating advanced 3D attention mechanisms into the neural ODE solver and enforcing strict hard physical constraints during sampling, our model successfully bridges isolated borehole sequences to reconstruct globally coherent strata. Extensive Out-Of-Distribution (OOD) evaluations demonstrate that 3D-GeoFlow achieves a paradigm shift over both heuristic interpolations and standard generative baselines, yielding unprecedented geological realism and structural fidelity. Future work will integrate multi-modal geophysical responses, including gravity and magnetic fields, as dense soft constraints to better resolve micro-scale, long-tail anomalies missed by sparse drilling.

\bibliographystyle{plain} 
\bibliography{references}   





\end{document}